\documentclass[runningheads]{llncs}
\usepackage{graphicx}
\usepackage[utf8]{inputenc}
\usepackage{amsmath}
\usepackage{mathtools}
\usepackage{amssymb}
\usepackage{graphicx}
\usepackage{algorithm2e}
\usepackage{multirow}
\usepackage{hyperref}
\usepackage{pifont}
\usepackage{color}
\usepackage[table]{xcolor}
\usepackage{hyperref}
\usepackage[capitalize]{cleveref}
\usepackage{enumitem}
\usepackage{array}
\usepackage[font=footnotesize,skip=0pt]{caption}
\usepackage{graphicx}
\usepackage[export]{adjustbox}
\usepackage{multirow}
\usepackage{makecell}
\usepackage{siunitx}
\usepackage{float} 
\usepackage[usestackEOL]{stackengine}
\setstackgap{L}{-3em}
\usepackage{subfig}
\usepackage{hhline}
\usepackage[super]{nth}
\usepackage[autostyle=true]{csquotes}

\usepackage[capitalize]{cleveref}
\crefname{section}{Sec.}{Secs.}
\crefname{subsection}{Sec.}{Secs.}
\crefname{subsubsection}{Sec.}{Secs.}
\crefname{table}{Tab.}{Tabs.}
\crefname{figure}{Fig.}{Figs.}
\crefname{algorithm}{Alg.}{Algs.}

\newcolumntype{L}[1]{>{\raggedright\let\newline\\\arraybackslash\hspace{0pt}}m{#1}}
\newcolumntype{C}[1]{>{\centering\let\newline\\\arraybackslash\hspace{0pt}}m{#1}}
\newcolumntype{R}[1]{>{\raggedleft\let\newline\\\arraybackslash\hspace{0pt}}m{#1}}

\newcommand{\cmark}{\ding{51}}%
\newcommand{\xmark}{\ding{55}}%

\newcommand{\p}[1]{\phantom{#1}}

\newcommand{\inlinefig}[1]{%
  \begingroup\normalfont
  \includegraphics[height=\fontcharht\font`\B]{#1}%
  \endgroup
}

\newcommand{\iconc}[1]{\inlinefig{figures/cluster_icon/c#1}}     
\newcommand{\iconl}[1]{\inlinefig{figures/label_icon/l#1}}       
\newcommand{\icono}[1]{\inlinefig{figures/outlier_icon/o#1}}     
\newcommand{\iconp}[1]{\inlinefig{figures/protocol_icon/p#1}}    
\newcommand{\iconh}[1]{\inlinefig{figures/locality_icon/h#1}}    
\newcommand{\icona}[1]{\inlinefig{figures/application_icon/a#1}} 
\newcommand{\icone}[1]{\inlinefig{figures/experiment_icon/e#1}}  

\newcommand\hfilll{\hspace{0pt plus 1filll}}
\begin{document}
\renewcommand{\arraystretch}{.9}
\title{A Study of Deep Learning for \\ Network Traffic Data Forecasting} 
%
%
\author{Benedikt Pfülb\inst{1},
Christoph Hardegen\inst{1}, \\
Alexander Gepperth\inst{1} and
Sebastian Rieger\inst{1}}
\authorrunning{B.\ Pfülb et al.}
%
\institute{University of Applied Sciences Fulda, Leipziger Straße 123, 36037 Fulda, Germany
\url{http://www.hs-fulda.de} \\
\email{\{benedikt.pfuelb,christoph.hardegen,alexander.gepperth,sebastian.rieger\} \\ @cs.hs-fulda.de}}
\maketitle              
\begin{abstract} 
We present a study of deep learning applied to the domain of network traffic data forecasting.
This is a very important ingredient for network traffic engineering, e.g., intelligent routing, which can optimize network performance, especially in large networks.
In a nutshell, we wish to predict, in advance, the bit rate for a transmission, based on low-dimensional connection metadata (\enquote{flows}) that is available whenever a communication is initiated.
Our study has several genuinely new points:
First, it is performed on a large dataset (\num{\approx 50} million flows), which requires a new training scheme that operates on successive blocks of data since the whole dataset is too large for in-memory processing.
Additionally, we are the first to propose and perform a more fine-grained prediction that distinguishes between low, medium and high bit rates instead of just \enquote{mice} and \enquote{elephant} flows.
Lastly, we apply state-of-the-art visualization and clustering techniques to flow data and show that visualizations are insightful despite the heterogeneous and non-metric nature of the data.
We developed a processing pipeline to handle the highly non-trivial acquisition process and allow for proper data preprocessing to be able to apply DNNs to network traffic data. 
We conduct DNN hyper-parameter optimization as well as feature selection experiments, which clearly show that fine-grained network traffic forecasting is feasible, and that domain-dependent data enrichment and augmentation strategies can improve results.
An outlook about the fundamental challenges presented by network traffic analysis (high data throughput, unbalanced and dynamic classes, changing statistics, outlier detection) concludes the article.
\keywords{DNN \and Incremental Learning \and Network Traffic Engineering.}
\end{abstract}
\section{Introduction}\label{sec:introduction}
\begin{figure}
  \centering
  \includegraphics[width=\textwidth, trim= -.5cm -.25cm -.5cm 0cm]{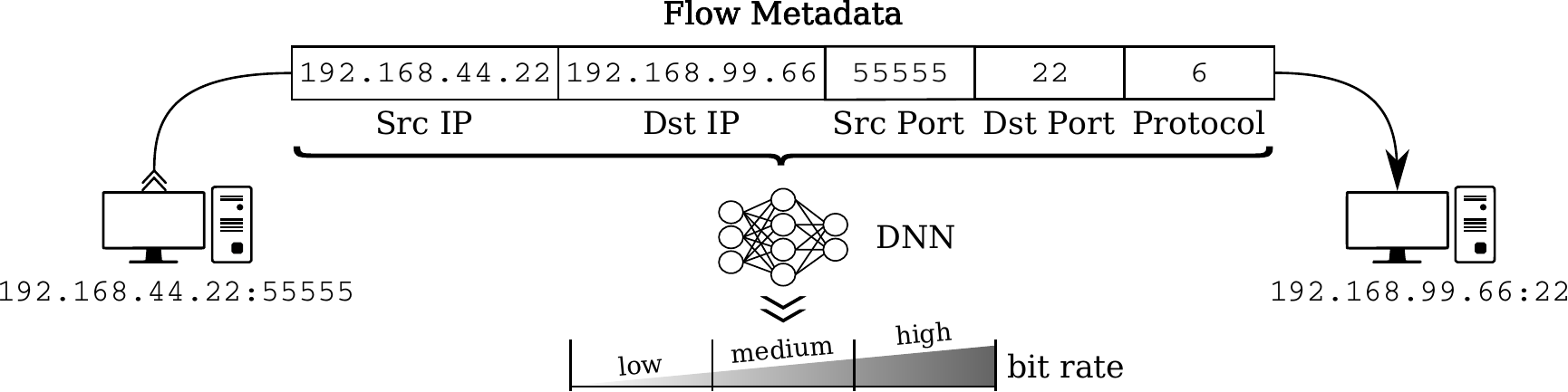}
  \caption{\label{fig:intro}
    Overview of the principal task of network traffic forecasting.
    Upon establishment of an IP-based network communication between two computers, the metadata (5-tuple) is supplied to a trained DNN that forecasts the bit rate (low, medium and high) for this communication.
    This is done before any data is exchanged.
    In order to train the DNN, target values have to be obtained after a communication is terminated.
  }
\end{figure}

This article is in the context of computer network traffic forecasting.
We focus on using deep neural networks (DNNs).
More precisely, we investigate how DNNs can predict, in advance, the approximate bit rate of a computer network communication.
This is modeled as a classification task with three classes (low, medium and high).
The key idea here is to take this decision based only on the metadata of the communication, which are represented, in their most basic form, as a 5-tuple: source and destination IP address, source and destination port as well as the transport protocol, e.g., TCP or UDP.
An example of the flow metadata as well as the classification task is depicted in \cref{fig:intro}.
%
%
\par
The motivation for investigating this kind of classification problem stems from the field of software-defined networking (SDN).
While traditional and still most prevalent network routing algorithms are primarily based on the destination address, SDN-like techniques enable dynamic determination of paths based on traffic characteristics.
For example, routers can typically choose between several paths to forward network traffic to a specific destination.
On the one hand, in the case of paths with unequal costs, using only the optimal path could cause congestion while alternative paths are underutilized.
On the other hand, using the hash of a 5-tuple to decide between multiple equal-cost paths might lead to unequal load balancing because the amount of transmitted data cannot be considered in advance.
Also, the path cannot easily be changed during the communication.
Therefore, predicting the bit rate of a communication beforehand is of high value for the routing and load balancing process.
%
\subsection{Problem Formulation and Approach of the Article}\label{sec:motivation}
\textbf{Challenges}
The principal immediate challenges for machine learning in network traffic forecasting raised and addressed in this article are as follows:
\begin{itemize} 
\item data acquisition: Here, one encounters difficulties creating the technical infrastructure (i.e., administrative access to network devices, handling large amounts of data resulting from capturing the network traffic) and the fact that metadata contain sensitive information, requiring an anonymization strategy that preserves information content and relations.
Furthermore, the encoding of metadata into a form that is suitable for DNNs and the generation of target values is essential.
\item regression problem: Network traffic forecasting is essentially a regression problem as a continuous and highly variable quantity (the bit rate) needs to be predicted, which is a challenging task that must be simplified suitably.
\item class imbalance: Communications transmitting very few data are much more frequent than those transferring huge amounts of data\cite{nguyen2008survey}.
The distribution regarding the bit rate as target value can be expected to change over time.
\item concept drift: The statistics of the problem may be time-dependent, e.g., depending on the day of week, the time of day, the season, technical changes, etc.
A DNN classifier trained on day $X$ may therefore not be suited to classify metadata collected on day $Y\neq X$.
We are therefore dealing with a problem where continual re-training must be conducted while retaining previous knowledge (see \cite{pfuelb2019a} for a recent review on this kind of training paradigm).
\item big data setting: The amount of flows is so high, and their variability so significant, that DNN training on a representative training set can no longer be performed in-memory.
In our scenario, the network devices we accessed to collect data delivered \num{57} million records in \num{8} hours (about \SI{15}{\giga\byte} of raw data respectively $\varnothing$ $\approx$\num{2000} flows per second, including the 5-tuple).
\end{itemize}
\smallskip
\par
\noindent \textbf{Approach} In order to address these challenges, we first of all treat DNN training as a streaming problem by dividing all collected metadata into blocks of \num{100000} flows each.
Training and evaluation are then conduced in a semi-streaming fashion, starting with the first block and subsequently passing to following ones, with all relevant preprocessing operations being performed block-wise.
Concept drift is thus incorporated into DNN training although it cannot be completely compensated.
The class imbalance problem is currently fixed by different class balancing mechanisms, since the whole reference dataset\footnote{Our anonymized dataset is available upon request.} is known prior to DNN training.
This will have to be replaced by more generic solutions in the future.
Lastly, we transform the regression problem into a classification problem with three classes, thus balancing the need for precision and complexity of the \mbox{problem}.
\subsection{Related Work}\label{sec:related_work}
Network traffic forecasting with machine learning techniques is a field (see \cite{nguyen2008survey} for a review) that is receiving increased attention, probably due to the recent advances in machine learning techniques, notably deep learning models.
From a machine learning point of view, many recent articles can be grouped according to whether they conduct online or offline learning on streaming network data, what machine learning models they employ in general, what dataset they operate on and whether they systematically investigate the effects of data enrichment.
To the best of our knowledge, all related works operate on datasets of around \num{1000000} flows which is significantly smaller than the dataset we use in this study, and thereby avoid \enquote{big data} issues like the necessity to perform learning in blocks.
Furthermore, related works reduce the network traffic forecasting problem to a binary classification into \enquote{mice} and \enquote{elephant} flows.
\par
In \cite{poupart2016online}, the authors apply online and offline learning methods (Multi-Layer Perceptron, Gaussian Process Regression and Online Bayesian Moment Matching).
The problem is treated as a two-class classification problem using three different datasets, one self-created (not available) and the others from other authors \cite{benson2010network}.
No data enrichment is performed, however information about the first three exchanged packets is used in addition to a flow's 5-tuple as a basis for classification, which differs from our approach that does not consider such information.
In \cite{xiao2015efficient}, purely offline learning with two-class decision tree classifiers is performed on the \enquote{Wide} dataset and a self-created one (not available) coming from a data center, also without data enrichment.
In \cite{wang2016framework}, semi-supervised SVMs are trained in an offline fashion to solve a two-class problem using a simple form of data enrichment.
Evaluations are conducted on a dataset of approximately \num{1000000} flows, \enquote{captured by the Broadband Communication Research Group in UPC, Barcelona, Spain} (no reference given, no data available).
\cite{SHI20171} use offline SVM training on two datasets captured on Chinese university campuses (no reference given, not available), and experiment extensively with feature selection schemes, however based only on the basic 5-tuple information.
Another interesting albeit not directly related application of machine learning is the routing of flows itself (see \cite{Valadarsky:2017:LR:3152434.3152441}).
\subsection{Contribution of the Article}
Overall, this study shows that fine-grained network traffic forecasting using three classes with DNNs is feasible, and that it can be performed in a \enquote{big data} setting, operating on separate data blocks sequentially.
We furthermore investigate the effects of data enrichment beyond the basic 5-tuple information, while also dealing with anonymization and privacy issues.
Lastly, we show that modern data visualization and clustering techniques can be readily applied to network traffic data in order to gain deeper insights into the structure of the problems and to \enquote{debug} machine learning solutions.

\section{Flow Data Pipeline}\label{sec:data_pipeline}
\begin{figure}[H]
  \centering
  \includegraphics[width=\textwidth, trim=-.5cm -.25cm -.5cm 0cm, clip]{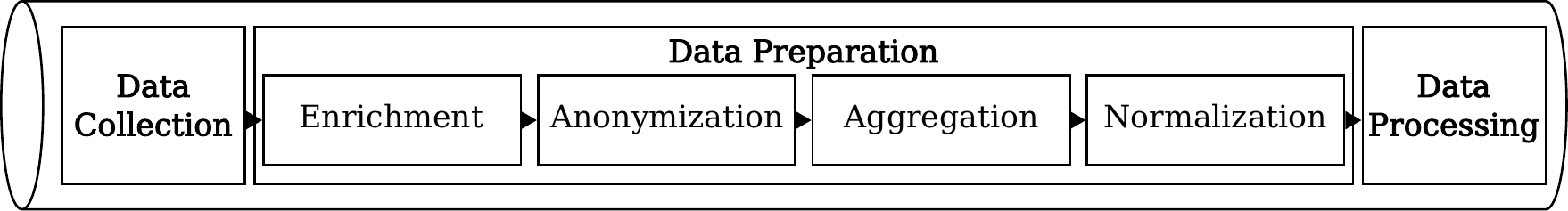}
  \caption{
    Overview of the stages of the flow data pipeline: data collection, preparation and processing.
    At first, flow records are collected.
    Before IP addresses are anonymized, further related metadata is added during the enrichment phase.
    Afterwards, the fusion of individual \textit{flow records} is applied to aggregate \textit{flow entries}.
    Flow data is normalized and stored as a dataset that is used for DNN training in the data processing phase.
  }
  \label{fig:flowdatapipeline}
\end{figure}

We introduce a flow data pipeline (see \cref{fig:flowdatapipeline}) that is responsible for collecting the network traffic flows and producing a dataset consisting of flows describing communications.
Data collection and the first parts of the data preparation (enrichment and anonymization) are entirely performed within our data center to ensure privacy (supported by the administration).
The codebase of the pipeline is publicly available in our repository\footnote{\url{https://gitlab.informatik.hs-fulda.de/flow-data-ml}}.
\subsection{Data Collection}\label{sec:data_collection}
%
A flow is understood to be the history of a single transmission between two endpoints, from establishment to termination (only metadata).
In particular, flows are partly characterized by the 5-tuple.
Flows may include additional metadata, e.g., the duration or number of transferred bytes.
\par
Flow data is collected from the networks at Fulda University of Applied Sciences.
We export network flow data (\num{57} million flow records) using the NetFlow standard from the two core network devices in our university data center during a continuous time interval of \num{8} hours on a weekday (02/15/2019 9:00\,AM to 5:00\,PM).
These core components connect multiple subnets from the data center, laboratory, WiFi and campus networks.
Collecting data from these diverse networks ensures realistic traffic characteristics and patterns to be used for the subsequent data analysis and network flow prediction.
For example, collected traffic patterns include internal and external flows originating from client-to-server as well as server-to-server communication.
\subsection{Data Preparation}\label{sec:data_preparation}
Due to the extremely large amount of collected flow records, these are partitioned into separate \textit{blocks} of \num{100000} records each (representing $\varnothing$ $\approx$\SI{1}{\minute}), and the operations given below are applied block-wise.
We thus obtain the final dataset, which is used for all experiments in this article, containing about \num{53} million aggregated \textit{flow entries} out of approximately \num{57} million captured \textit{flow records}.
\smallskip
\par
\noindent \textbf{Enrichment}\label{sec:enrichment}
Based on the collected 5-tuples, additional context information is derived.
For example, groups of internal and external addresses (e.g., IP subnets, VLANs and geographical regions) can be identified from the network addresses.
These contexts deliver additional characteristics and patterns for the subsequent analysis and the prediction process.
\smallskip
\par
\noindent \textbf{Anonymization}\label{sec:anonymisation}
To ensure that collected metadata cannot be traced back to individual network addresses and end users, while still keeping the syntax and semantic of the data intact to prevent distortion of contained characteristics for the subsequent analysis, an appropriate anonymization algorithm was developed.
This mechanism anonymizes all address-related metadata, i.e., IP and network addresses.
The data center that exports the traffic metadata defines a password, which is cryptographically hashed and used as a seed for randomized permutation tables.
A seed ensures consistent anonymization for further data acquisitions.
Each octet of an IP address is anonymized individually using these tables.
This way, the semantics of an address, e.g., regarding the relevance and order of the octets forming a group of network addresses, will stay intact after the anonymization and can still be used as a characteristic feature for prediction.
However, adjacency of addresses will not be preserved in favor of the anonymization due to the seeded randomization of the permutation tables.
\smallskip
\par
\noindent \textbf{Aggregation}\label{sec:aggregation}
Exported unidirectional \textit{flow records} that potentially represent only a part of a communication (due to exporter timeouts or cache sizes) are aggregated to ensure coherent \textit{flow entries}.
The aggregation of records is based on the 5-tuple and additional traffic characteristics, e.g., flags and predefined time intervals.
Duplicated flow records from both exporting network devices are filtered.
During this phase, the number of records is reduced to, on average, \SI{7.5}{\percent} of the collected flow records.
Afterwards, ports greater than \num{32767} are replaced by zero because they are chosen randomly by common operating systems.
\smallskip
\par
\noindent \textbf{Normalization}\label{sec:normalization}
We convert raw, heterogeneous features into a format suited for DNNs, e.g., a sequence of floating points, in three different ways:
Bit patterns are converted by promoting each bit to a \num{0.0} or \num{1.0}, float values are interpolated between $0.0$ and $1.0$ (min-max normalization) and categorical values are encoded as \enquote{one-hot} vectors, i.e., a single value of \num{1.0} put at an unique position, having a length of $N$, where $N$ represents the number of distinct categories.
An example is given in \cref{tab:normalization_example}.
\begin{table}
  \centering
  \caption{
    Exemplary normalization of an IP address, a port and a protocol value using different data formats (bit pattern or float value).
    Each data type has a feature-dependent size specifying the number of individual float values that are used as input for the DNN.
    For example, next to its raw format, each octet of an IP address is represented in its original format as bit pattern or as float values.
  }
  \label{tab:normalization_example}
  \begin{tabular}{|l|l|L{4.8cm}|L{3.25cm}|}
    \hline
    \textbf{Feature} & \textbf{Raw format} & \textbf{Bit pattern} (size)                                           & \textbf{Float value(s)} (size)                  \\ \hline
    IP address       & 81.169.238.182    & 0,1,0,1,0,0,0,1,1,0,1,0,1,0,0,1, 1,1,1,0,1,1,1,0,1,0,1,1,0,1,1,0 (32) & \makecell[l]{0.3176, 0.6627,\\0.9333, 0.7137 (4)} \\ \hline
    Port             & 80                & 0,0,0,0,0,0,0,0,0,1,0,1,0,0,0,0 (16)                                  & 0.0012 (1)                                     \\ \hline
    Protocol         & 6                 & 0,0,0,0,0,1,1,0 (8)                                                   & 0.0235 (1)                                     \\ \hline
  \end{tabular}
\end{table}
\par
The output of the normalization and thus of the data preparation process is the actual dataset (about \SI{2.3}{\giga\byte}).
Next to the bit rate, there are other flow features that can be used as class labels and hence for a prediction, e.g., the number of transferred bytes or the duration of a flow.
A combination of selected labels is conceivable as well.
The datasets structure is summarized in \cref{tab:dataset_features}.
\begin{table}
  \centering
  \caption{
    Overview of the dataset features and labels.
    For each raw flow feature, the supported respectively used (gray highlighting) data formats are shown, and the number of values is given, as well as the point in the flow data pipeline in which the information is added (Src).
    Features for both source and destination are marked with $\rightleftarrows$.
  }
  \label{tab:dataset_features}
  \begin{tabular}{|l|C{0.8cm}|C{0.6cm}|C{0.8cm}|C{0.6cm}|C{1pt}|l|C{0.8cm}|C{0.6cm}|C{0.8cm}|C{0.6cm}|}
    \cline{1-5}\cline{7-11}
    \multirow{2}{*}{\textbf{Feature}}            &       \multicolumn{3}{c|}{\textbf{Data format}}       & \multirow{2}{*}{\textbf{Src}} && \multirow{2}{*}{\textbf{Feature}}              &                \multicolumn{3}{c|}{\textbf{Data format}}                 & \multirow{2}{*}{\textbf{Src}} \\ \cline{2-4}\cline{8-10}
                                                 &        Float         &          Bit          &   OH   &                               &&                                                &        Float         &          Bit          &            OH             &                               \\ \hhline{|-|-|-|-|-|~|-|-|-|-|-|}
    \verb|month|                                 & \cellcolor{gray!25}1 &           4           &   12   &    \multirow{8}{*}{DC}        && \verb|longitude| \hfilll $\rightleftarrows$    & \cellcolor{gray!25}1 &        \xmark         &          \xmark           &     \multirow{6}{*}{DE}       \\ \hhline{|-|-|-|-|~|~|-|-|-|-|~|}
    \verb|day|                                   & \cellcolor{gray!25}1 &           5           &   31   &                               && \verb|latitude| \hfilll $\rightleftarrows$     & \cellcolor{gray!25}1 &        \xmark         &          \xmark           &                               \\ \hhline{|-|-|-|-|~|~|-|-|-|-|~|}
    \verb|hour|                                  & \cellcolor{gray!25}1 &           5           &   24   &                               && \verb|country code| \hfilll $\rightleftarrows$ & \cellcolor{gray!25}1 &           8           &            240            &                               \\ \hhline{|-|-|-|-|~|~|-|-|-|-|~|}
    \verb|minute|                                & \cellcolor{gray!25}1 &           6           &   60   &                               && \verb|vlan| \hfilll $\rightleftarrows$         &          1           & \cellcolor{gray!25}12 &          \xmark           &                               \\ \hhline{|-|-|-|-|~|~|-|-|-|-|~|}
    \verb|second|                                & \cellcolor{gray!25}1 &           6           &   60   &                               && \verb|locality| \hfilll $\rightleftarrows$     &        \xmark        & \cellcolor{gray!25}1  &             2             &                               \\ \hhline{|-|-|-|-|~|~|-|-|-|-|~|}
    \verb|protocol|                              &          1           & \cellcolor{gray!25}8  & \xmark &                               && \verb|flags|                                   &          1           & \cellcolor{gray!25}8  &          \xmark           &                               \\ \hhline{|-|-|-|-|~|~|-|-|-|-|-|}
    \verb|address| \hfilll $\rightleftarrows$    &          4           & \cellcolor{gray!25}32 & \xmark &                               &&                                                                   \multicolumn{5}{c|}{}                                                                   \\ \hhline{|-|-|-|-|~|~|-|-|-|-|-|}
    \verb|port| \hfilll $\rightleftarrows$       &          1           & \cellcolor{gray!25}16 & \xmark &                               && \textbf{Label}                                 &                \multicolumn{3}{c|}{\textbf{Data format}}                 &         \textbf{Src}          \\ \hhline{|-|-|-|-|-|~|-|-|-|-|-|}
    \verb|network| \hfilll $\rightleftarrows$    &          4           & \cellcolor{gray!25}32 & \xmark &    \multirow{3}{*}{DE}        && \verb|duration|                                &        \xmark        &        \xmark         &          \cmark           &     \multirow{3}{*}{DA}       \\ \hhline{|-|-|-|-|~|~|-|-|-|-|~|}
    \verb|prefix length| \hfilll $\rightleftarrows$ &          1           & \cellcolor{gray!25}5  & \xmark &                               && \verb|bytes|                                   &        \xmark        &        \xmark         &          \cmark           &                               \\ \hhline{|-|-|-|-|~|~|-|-|-|-|~|}
    \verb|asn| \hfilll $\rightleftarrows$        &          1           & \cellcolor{gray!25}16 & \xmark &                               && \verb|bit rate|                                &        \xmark        &        \xmark         & \cellcolor{gray!25}\cmark &                               \\ \hhline{|-|-|-|-|-|~|-|-|-|-|-|}
                                                                                                                                         \multicolumn{11}{c}{}                                                                                                                                        \\[-2ex]
                                                                                           \multicolumn{11}{c}{\scriptsize{DC = Data Collection; DE = Data Enrichment; DA = Data Aggregation; OH = One-Hot}}
  \end{tabular}
\end{table}
\subsection{Data Processing}\label{sec:processing}
In the data processing phase a fully-connected DNN is trained to predict the bit rate of a communication.
During the processing of the created flow dataset, three steps are performed blockwise:
At first, a sub-dataset can be extracted by feature selection.
Afterwards, data samples are labeled based on predefined class boundaries, which are selected to fit an almost balanced data distribution (presented in \cref{sec:distribution}).
Finally, training and testing is done on each individual block sequentially.
To evaluate different hyper-parameter setups, we do a parameter optimization.
The detailed process and related results are presented in \cref{sec:network_flow_prediction}.
\section{Exploratory Data Analysis and Visualization}\label{sec:analysis}
To provide a better understanding of flow data, we explore the distribution of features used for labeling (see \cref{sec:distribution}) and visualize the intrinsic structure of the data (see \cref{sec:structural_context}).
The analysis is performed on the first \num{1000} flow entries (including all features) that are selected from the shuffled test data of the first block.
Due to this, the same t-SNE output is used for all context-related taggings.
No significant deviations were observed when performing this analysis on other blocks (every \nth{50} block was compared).
All comparisons of the tagged t-SNE outputs are done by visual inspection.
\subsection{Label Distribution}\label{sec:distribution}
\begin{figure}
  \centering 
  \subfloat[][
  number of bytes
  ]
    {\includegraphics[width=0.31\textwidth,type=pdf,ext=.pdf,read=.pdf, trim = .3cm .3cm .3cm .3cm, clip]{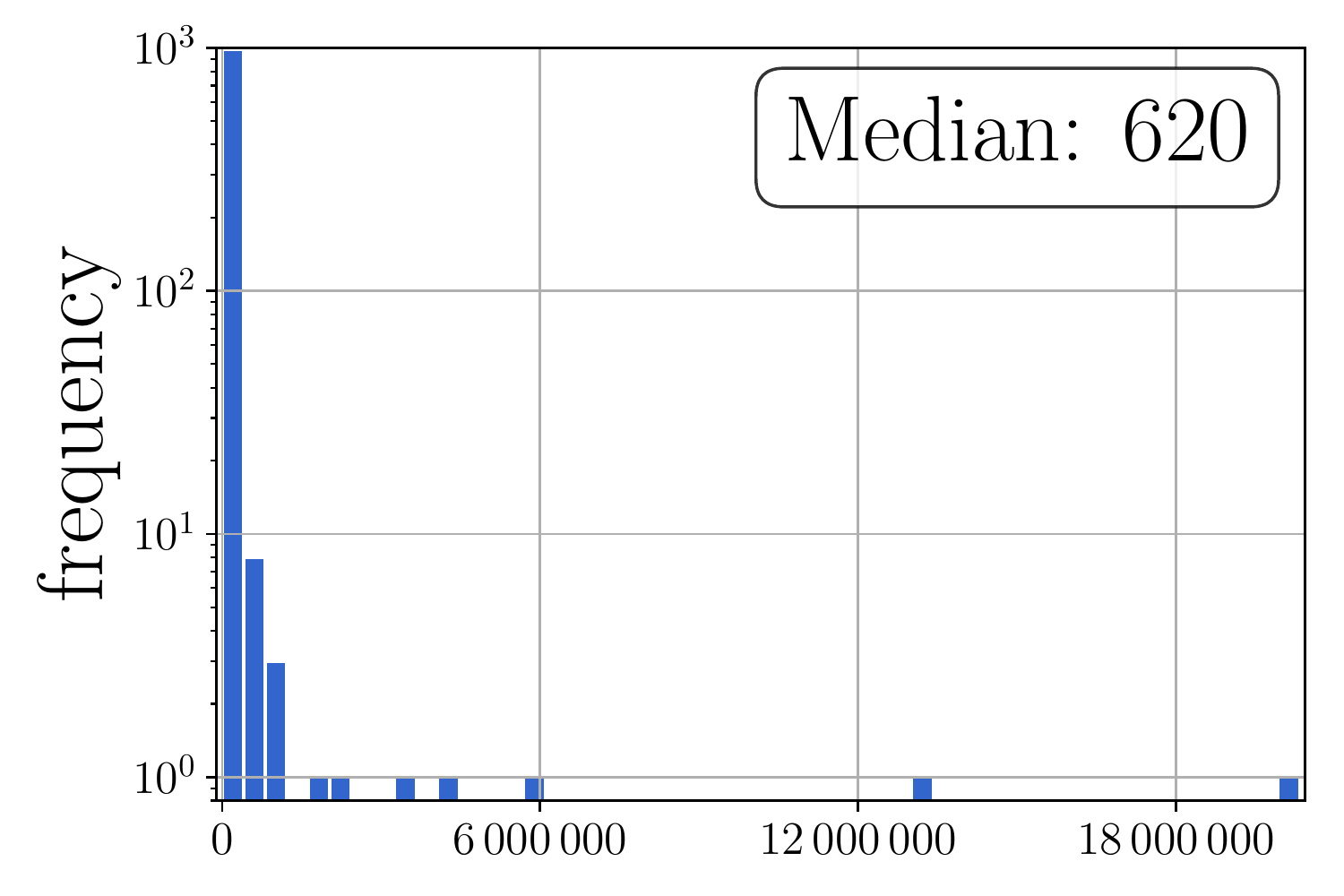}\label{fig:hist_bytes}} %
  \subfloat[][
    duration in $seconds$
    ]
    {\includegraphics[width=0.31\textwidth,type=pdf,ext=.pdf,read=.pdf, trim = .3cm .3cm .3cm .3cm, clip]{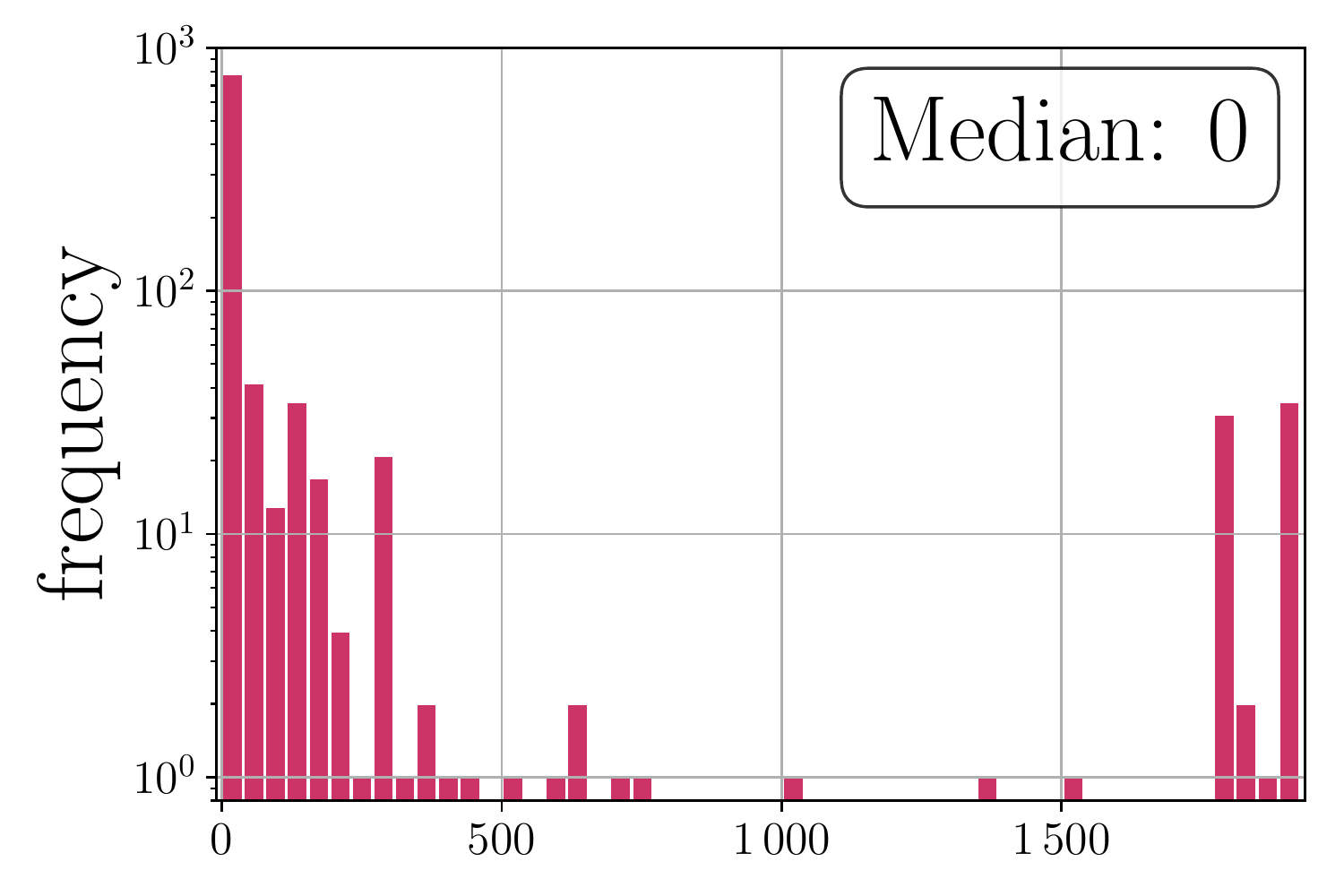}\label{fig:hist_duration}} %
  \subfloat[][
    bit rate in $^{bit}/_{sec}$
    ]
    {\includegraphics[width=0.31\textwidth,type=pdf,ext=.pdf,read=.pdf, trim = .3cm .3cm .3cm .3cm, clip]{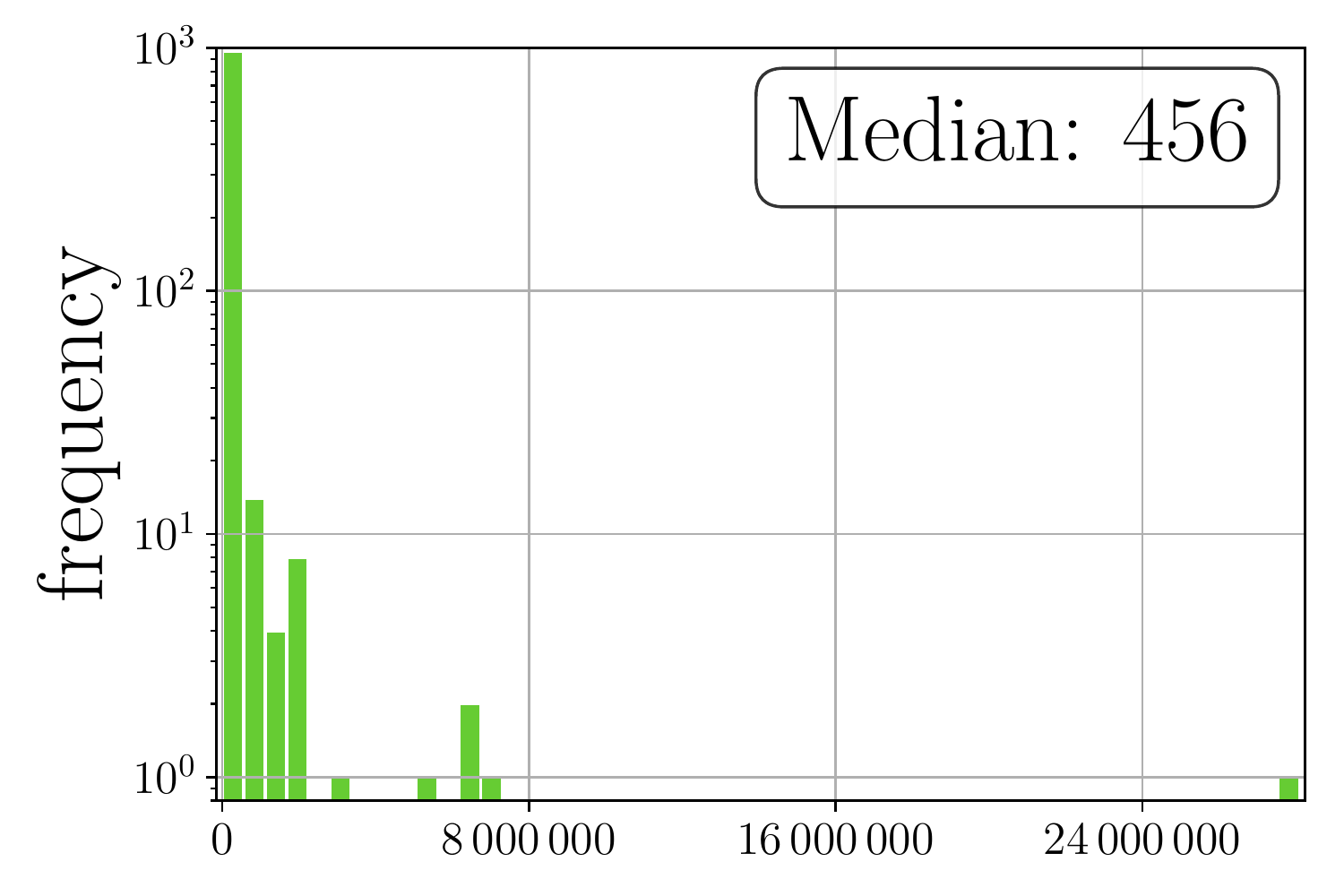}\label{fig:hist_bps}} %
  \vspace{0.2cm}
  \caption{
    Histograms for three possible flow labels of the selected \num{1000} elements in the first block.
    The majority of data samples have both a very small number of transferred bytes \protect\subref{fig:hist_bytes} and a short duration \protect\subref{fig:hist_duration}.
    Median values of \num{620} bytes respectively \num{0} seconds \mbox{(\SI{< 1000}{\ms})} substantiate this fact.
    Hence, the bit rate values \protect\subref{fig:hist_bps} are also very unevenly spread over the entire value range (median value is \num{456}).
  }
  \label{fig:label_histogram}
\end{figure}

We analyze the distribution of flow features that can be target values for traffic flow prediction, i.e., the transmitted bytes, the duration or the bit rate calculated from both.
Results are shown in \cref{fig:label_histogram}.
As other authors noted previously \cite{nguyen2008survey}, these features deviate strongly from a uniform distribution, which makes the determination of suitable class boundaries challenging.
The principal conclusion we draw from this is that we must use class balancing (see \cref{sec:network_flow_prediction}).
Although the data distribution justifies our class boundaries, their practical applicability, e.g., for intelligent routing, is questionable and considered as future work.
%
%
\subsection{Structural Context}\label{sec:structural_context}
\begin{figure}
  \centering
  \subfloat[][
  Tagging is based on the transport protocol of each flow entry.
  While the proportion of TCP is \SI{\approx 39}{\percent} (\num{389} flows, \iconp2), the one for UDP is \SI{\approx 60}{\percent} (\num{604} flows, \iconp1).
  There are separate spots for traffic data using either TCP or UDP.
  Besides TCP and UDP data, about \SI{1}{\percent} of the traffic data (\num{7} flows, \iconp3) is related to other protocols like ICMP.
  ]
  {\includegraphics[width=.9\textwidth,type=pdf,ext=.pdf,read=.pdf, trim = .3cm .3cm .3cm .3cm, clip]{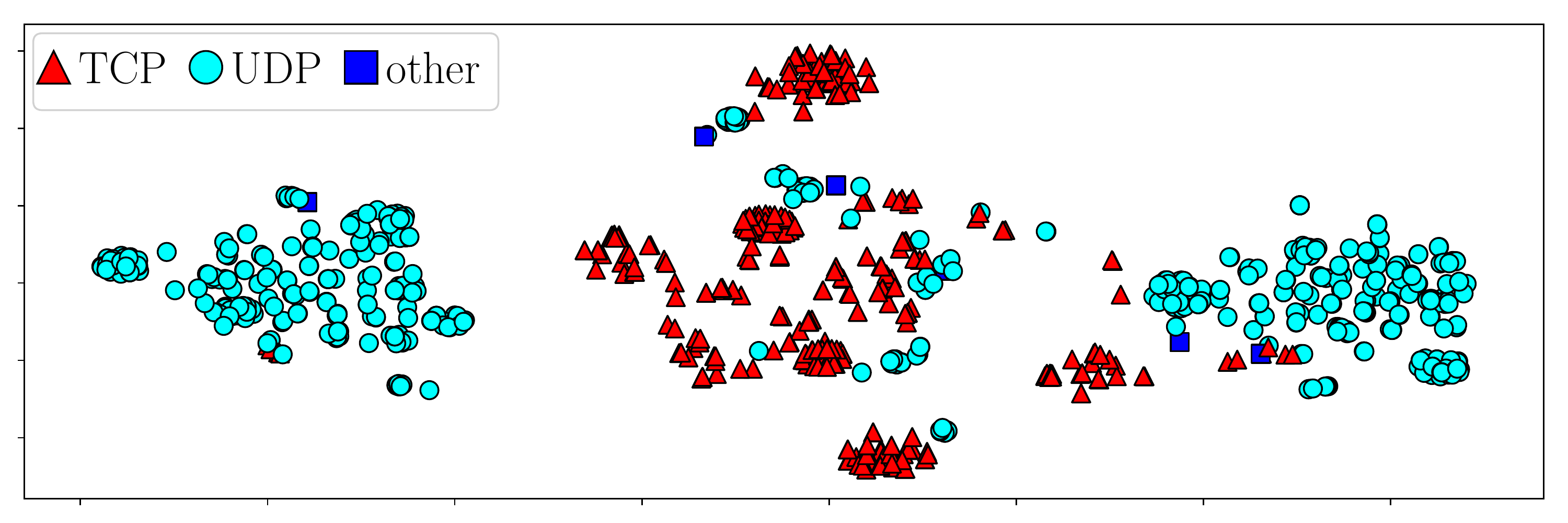}\label{fig:tsne_transport_protocol}} %
  \\ \vspace{-1em}
  \subfloat[][
  Tagging is based on the localities of each flow entry.
  Four different types are distinguished based on the combined locality of the source and destination system.
  Both can be either private or public.
  About \SI{92}{\percent} of the flows (\num{922}) describe traffic data where a public system is involved (\iconh1, \iconh2, \iconh3), while \SI{\approx 8}{\percent} (\num{78}) belong to communications between two private systems (\iconh4).
  Additionally, wireless (\SI{\approx 13}{\percent}, \num{133} flows) and wired network traffic (\SI{\approx 87}{\percent}, \num{867} flows) are separately delineated.
  According to \cref{fig:tsne_transport_protocol} and the shown locality, each part of the symmetric spots for WiFi traffic belongs to a specific transport protocol (TCP or UDP) and a separate communication direction.
  ]
  {\includegraphics[width=.9\textwidth,type=pdf,ext=.pdf,read=.pdf,trim = .3cm .3cm .3cm .3cm, clip]{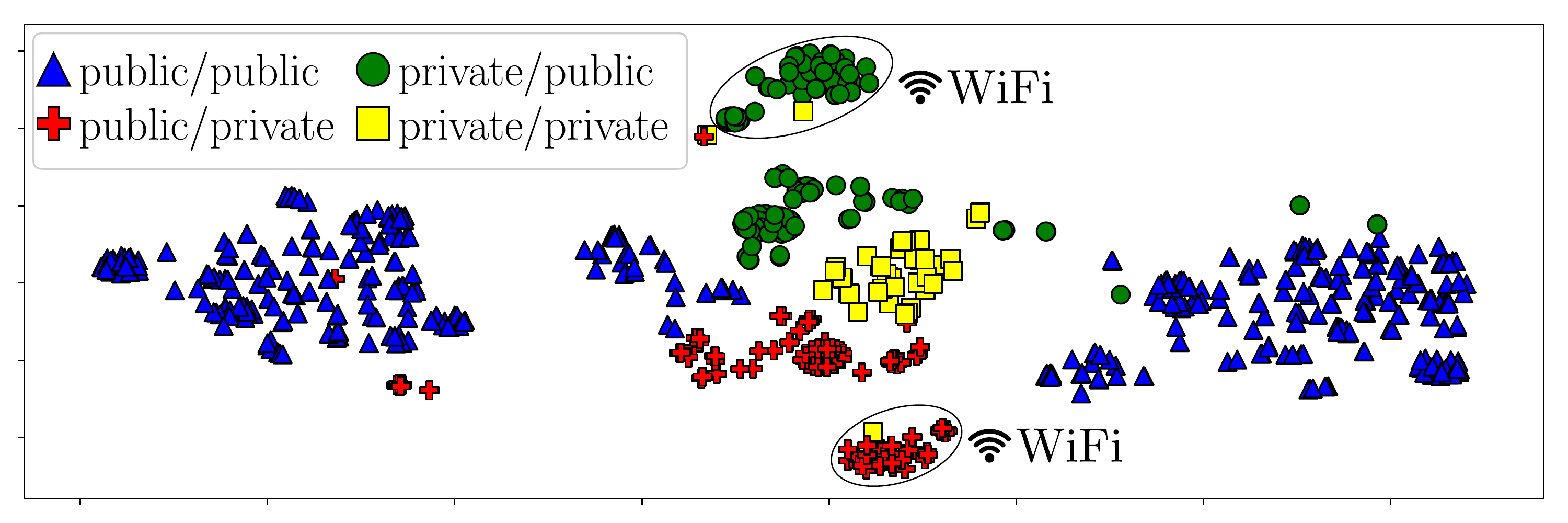}\label{fig:tsne_locality}} %
  \\ \vspace{-1em}
  \subfloat[][
  Tagging is based on the output of k-means clustering.
  Positions of cluster centers (\num{10}) are visualized according to t-SNE output.
  Nearly uniform accumulations of samples can be identified (e.g., \iconc3, \iconc8), clusters are spread (e.g.,~\iconc0,~\iconc9) and mixtures of samples of different clusters (e.g., \iconc2, \iconc4) are recognizable.
  ]
  {\includegraphics[width=.9\textwidth,type=pdf,ext=.pdf,read=.pdf, trim = .3cm .3cm .3cm .3cm,clip]{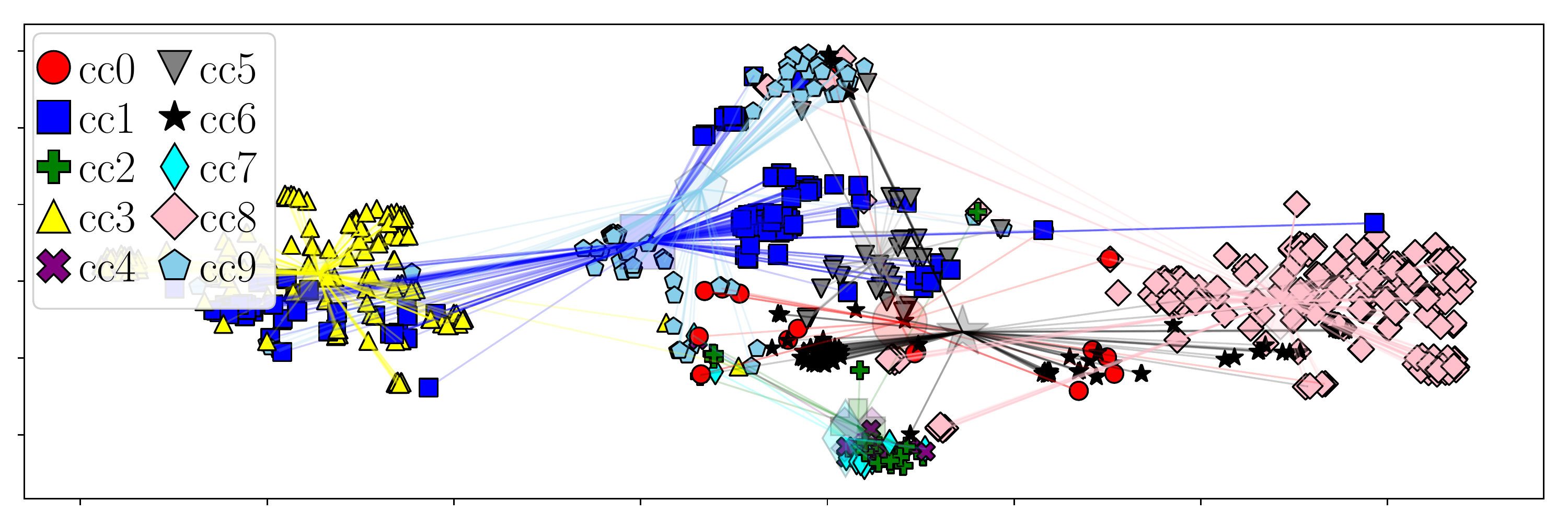}\label{fig:tsne_kmeans_centers}} %
  \vspace{0.2cm}
  \caption[]{
    Visualization of the selected \num{1000} flow samples.
    All figures show the same t-SNE results but tagging is based on the transport protocol \subref{fig:tsne_transport_protocol}, locality \subref{fig:tsne_locality} and clustering~\subref{fig:tsne_kmeans_centers}.
  }
  \label{fig:tsne_1}
\end{figure}

\begin{figure}[ht!]
  \centering
  \subfloat[][
  Tagging is based on an outlier detection using k-means clustering with 20 cluster centers.
  About \SI{11}{\percent} of the flow entries (\num{108}) are classified as outliers.
  Whereas outliers are marked with \icono2, kept data samples are shown as \icono1.
  ]
  {\includegraphics[width=.9\textwidth,type=pdf,ext=.pdf,read=.pdf, trim = .3cm .3cm .3cm .3cm, clip]{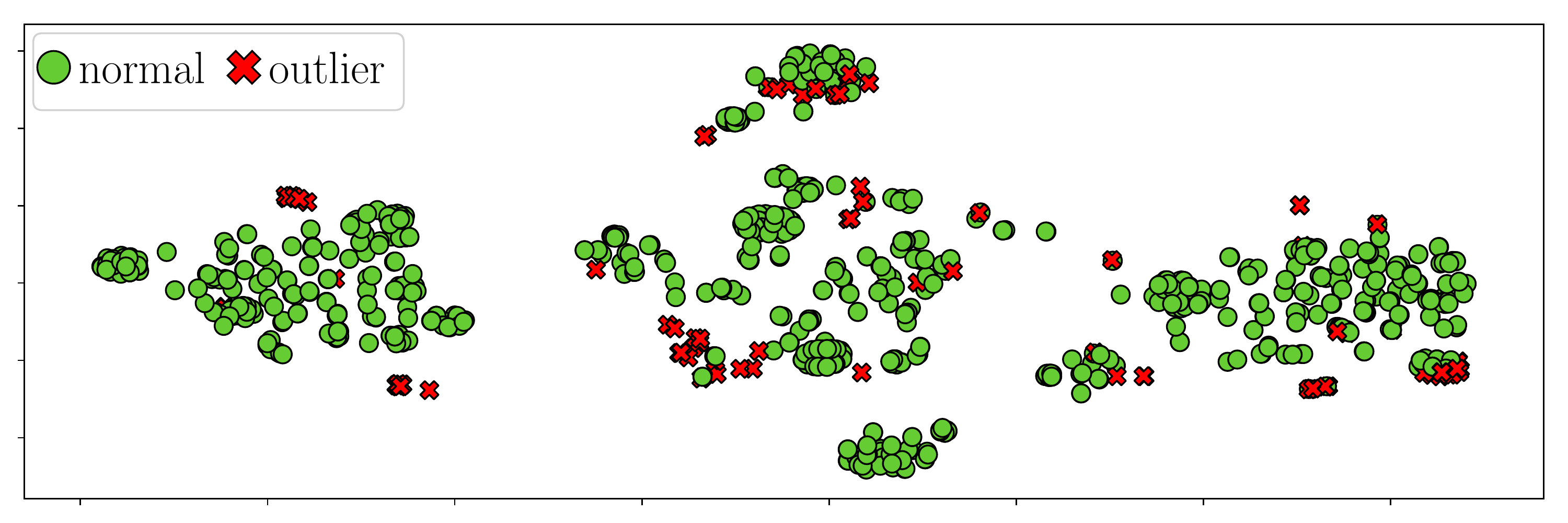}\label{fig:tsne_outlier}} %
  \\ \vspace{-1em}
  \subfloat[][
  Tagging is based on most frequent application protocols.
  Most samples belong to DNS communications (\SI{\approx 57}{\percent}, \num{571} flows, \icona1).
  Next to HTTP(S) (\SI{\approx 24}{\percent}, \num{241} flows, \icona0), other application protocols are visualized (\SI{\approx 19}{\percent}, \num{188} flows, \icona2).
  The latter include, for example, authentication, network monitoring and mail.
  ]
  {\includegraphics[width=.9\textwidth,type=pdf,ext=.pdf,read=.pdf, trim = .0cm .0cm .0cm .0cm, clip]{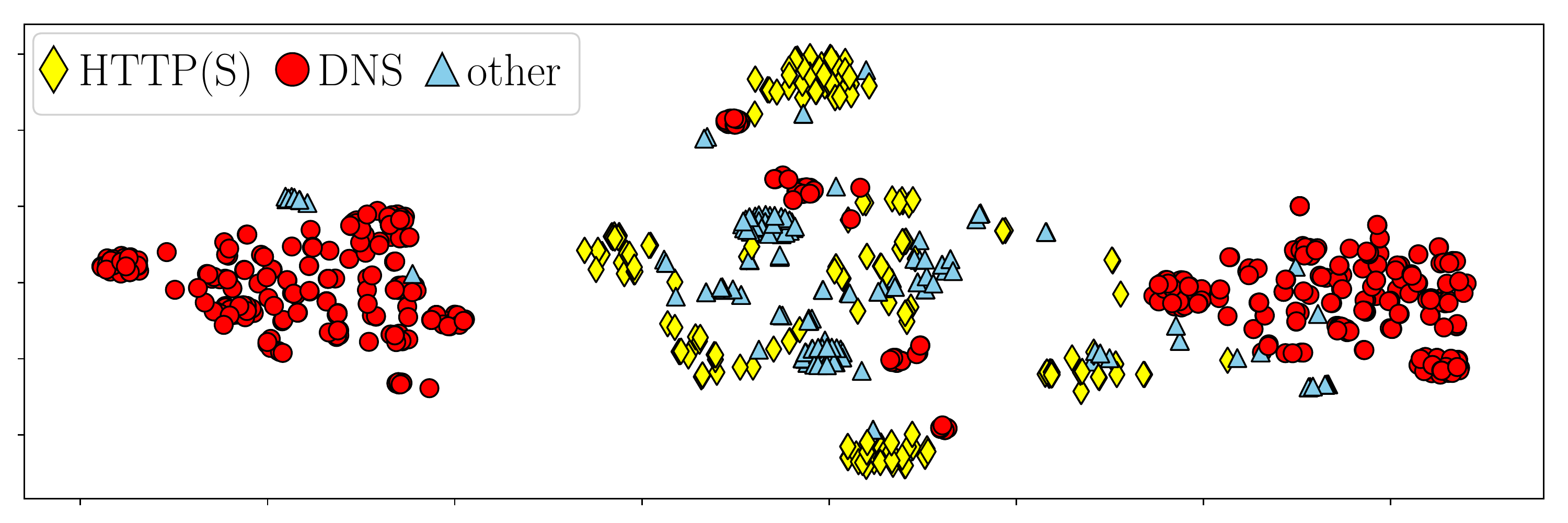}\label{fig:tsne_applications}} %
  \vspace{0.2cm}
  \caption[]{
    Visualization of structures in the selected \num{1000} flow samples.
    Tagging of the t-SNE result is based on an outlier detection \subref{fig:tsne_outlier} and the applications protocols \subref{fig:tsne_applications}.
  }
  \label{fig:tsne_2}
\end{figure}

To discover structural relations and similarities between individual flow entries (see \Cref{fig:tsne_1,fig:tsne_2}), we use t-Distributed Stochastic Neighbor Embedding (\mbox{t-SNE}) \cite{Maaten2008}, a state-of-the-art visualization method, which maps high-dimensional data samples to a low-dimensional space (2D or 3D).
We use the \mbox{t-SNE} implementation of the scikit-learn framework, parameter values being an iteration counter of \num{500}, a perplexity of \num{50} and a learning rate of \num{200}.
\par
\cref{fig:tsne_transport_protocol} illustrates feature similarities between flow entries that have a common transport protocol.
Two symmetric accumulations indicate opposite directions of the same communication.
Furthermore, there are examples that do not share the same transport protocol, but t-SNE points out similar feature data.
\par
Tagging of each data sample according to its type of communication, which is the combination of the source and destination locality (either private or public), is shown in \cref{fig:tsne_locality}.
For each communication type symmetric accumulations can be identified, whereas coherent spots map to individual flow directions.
\par
Additionally, we apply the k-means clustering algorithm on the sub-dataset and use the result for tagging the data samples in the t-SNE output.
With \mbox{k-means}, high-dimensional data samples are grouped around a predefined number of iteratively relocated cluster centers.
We use the implementation of tensorflow (v1.12) with \num{10} cluster centers, whereby the initial location of each center is determined randomly and the squared Euclidean distance is used as metric.
The tagging of the t-SNE output based on k-means clustering for the data samples is shown in \cref{fig:tsne_kmeans_centers}.
According to the t-SNE results, it can be observed that there are samples that belong to the same cluster but have certain feature differences and that there are samples of different clusters sharing feature properties.
The actual results depend on the chosen number of cluster centers.
\par
We also perform an outlier detection for each k-means cluster using different metric thresholds (average and median distance as well as both summed up with the standard deviation).
See \cref{fig:tsne_outlier} for an exemplary presentation of detected outliers.
With regard to our experiments described in the next section, the outlier detection has no significant influence on network flow prediction.
\par
According to \cref{fig:tsne_applications} DNS and HTTP(S) are the most used application protocols in the dataset.
The huge proportion of DNS traffic states the rate of flow entries with a low bit rate respectively short duration. \\
\par
The data analysis emphasizes relations and feature similarities between individual data samples.
All visualizations use the same t-SNE output, but context-related tagging, e.g., regarding used protocols or communication directions, helps to clarify different structural patterns within network flow data.
%
\section{Network Flow Prediction Experiments with DNNs}\label{sec:network_flow_prediction}
We employ a fully-connected DNN with $L$ layers of identical sizes $S$, each hidden layer applying a ReLU transfer function whereas the output layer applies a softmax function.
The batch size $bs$\,=\,\num{100} and the number of training epochs $\mathcal{E}$\,=\,\num{10} are fixed for all experiments.
DNN training minimizes a standard cross-entropy loss by stochastic gradient descent by means of the Adam Optimizer.
The last \SI{10}{\percent} of a chronologically ordered data block are completely used for testing every \nth{50} iteration.
\par
\noindent\textbf{Choice of evaluation metrics}
Since we are dealing with a three-class problem, the usual metrics for binary problems are not applicable, such as F1 score, precision, recall, etc. 
Instead, we present results in the more general form of a confusion matrix, from which we can derive classification accuracy by considering only the diagonal elements. 
Both of these measures are applicable for classification tasks with an arbitrary number of classes, which can be useful for comparison should we decide to introduce more classes at a later point. 
In order to allow a more in-depth comparison between the experimental conditions (using the 5-tuple information vs.\ using all features), we decided to additionally compute the standard binary performance metrics separately for each class.
\par
\noindent\textbf{Hyper-parameters}
Tunable parameters include the learning rate $\epsilon$ and the optional application of dropout to input $d_i$ and hidden layers $d_h$, with different dropout probabilities.
The assignment of labels is done based on a class boundary parameter $\mathcal{C}$.
This list of boundary values is consistently used for all blocks before a training phase.
In order to specify the class balancing method, the parameter $\mathcal{W}$ is introduced.
Balancing for training and test data is achieved either by standard class weighting or under-sampling. 
Furthermore, a feature selector $\mathbb{F}$ provides support for the construction of sub-datasets.
$O_c$ specifies the number of cluster centers that are used for outlier detection using k-means clustering.
All hyper-parameters mentioned here ($L$, $S$, $\epsilon$, $d_i$, $d_h$, $\mathcal{C}$, $\mathcal{W}$, $\mathbb{F}$, $O_c$) are varied to perform a joint parameter optimization.
\par
We train all DNN classifiers on the first \num{10} blocks sequentially and evaluate the achieved prediction accuracy on each block's test set.
In order to obtain the best possible results, we conduct a combinatorial hyper-parameter optimization, leading to a total of \num{5400} DNN training and evaluation runs.
The explored parameter ranges are summarized in \cref{tab:parameter_optimization}.
Depending on the hardware, the computation time of one experiment is between \num{8} and \num{15} minutes.
Based on the complexity of the DNN and the chosen parameters, the GPU memory usage is between \num{140} and \SI{264}{\mega\byte} and the RAM utilization varies from \num{762} to \SI{1200}{\mega\byte}.
\begin{table}[b!]
  \centering
  \caption{
    Overview of the variables and tested values for parameter optimization.
  }
  \label{tab:parameter_optimization}
  \begin{tabular}{|l|c|c|}
    \hline
    \textbf{Parameter}                  &       \textbf{Variable}        &             \textbf{Values}                                                             \\ \hline
    Dropout (input, hidden)             &          $(d_i, d_h)$          & $\{(1.0, 1.0), (0.9, 0.6), (0.8, 0.5)\}$                                                \\ \hline
    Layers                              &              $L$               &               $\{3,4,5\}$                                                               \\ \hline
    Neurons per layer                   &              $S$               &     $\{200, 400, 600, 800, 1\,000\}$                                                    \\ \hline
    Learning Rate                       &           $\epsilon$           &        $\{0.01, 0.001, 0.0001\}$                                                        \\ \hline
    Features                            &          $\mathbb{F}$          &     $\{5\text{-tuple}, \text{all}\}$                                                    \\ \hline
    Class boundaries (bit rate)         &        $\mathcal{C}$           &        \makecell{$\{\{0, 500, 5\,000, \infty\}$, \\ \p{ }$\{0, 50, 8\,000, \infty\}\}$} \\ \hline
    \multirow{2}{*}{Class balancing method}    & \multirow{2}{*}{$\mathcal{W}$} &         $\{0$ (under-sampling),                                                         \\
                                        &                                &        \p{ \,}$1$ (class weighting)$\}$                                                 \\ \hline
    Cluster centers (outlier detection) &            $O_c$               &          $\{0,20,60,100,500\}$                                                          \\ \hline
  \end{tabular}
\end{table}
\par
\noindent \textbf{Labeling}\label{sec:labelling}
Because of the unbalanced data distribution (see \cref{sec:distribution}) that makes regression problematic (also addressed in \cite{poupart2016online}), we treat network flow prediction as a classification problem, using the three exemplary classes \enquote{low}, \enquote{medium} and \enquote{high}.
The calculated bit rate of each flow is used for computing a class based on thresholding operation (with the two thresholds adapted such that the distribution of classes is approximately flat).
Next to the used set of boundaries for class division, \cref{tab:exemplary_classes} presents related characteristics for each class.
\begin{table}
  \centering
  \caption{
    Exemplary class partitioning for the prediction of a flow's bit rate.
    Next to the related intervals, the median and mean value, the average number of elements using class balancing (class weighting or under-sampling) and the data distribution within each class for the first \num{10} blocks of the dataset are shown (log scale).
  }
  \label{tab:exemplary_classes}
  \begin{tabular}{|C{1cm}|C{1.6cm}|C{1.7cm}|C{1.5cm}|C{1.5cm}|C{3.9cm}|}
    \hline
    \multirow{2}{*}{\textbf{Class}} & \multirow{2}{*}{\textbf{Interval}} &         \multirow{2}{*}{\textbf{Median/}}         & \multicolumn{2}{c|}{\textbf{Average elements}} &                                                    \multirow{2}{*}{\textbf{Data}}                                                     \\ \cline{4-5}
    &                                    &                   \textbf{Mean}                   & class weighting &        under-sampling        &                                                         \textbf{distribution}                                                         \\ \hline
    0                &              $[0,50)$              & \makecell{\p{242\,73}$0$/ \\ \p{242\,73}$2$\p{/}} &    $21\,945$    &                              & \raisebox{-.5\height}{\includegraphics[height=2em, width=\linewidth, type=pdf,ext=.pdf,read=.pdf]{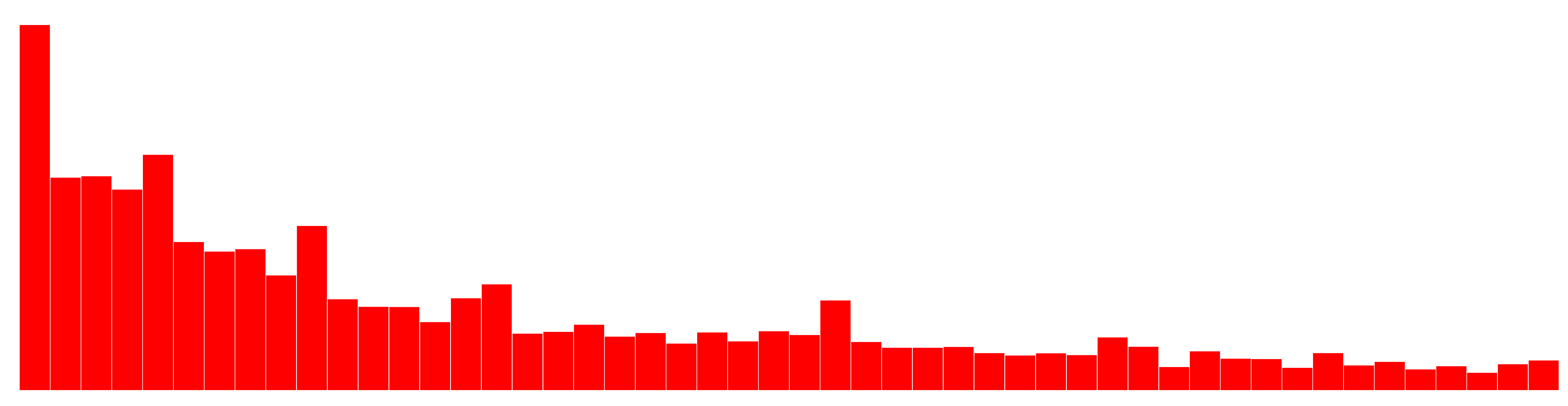}}  \\ \cline{0-3}\cline{6-6}
    1                &           $[50,8\,000)$            & \makecell{\p{24}$3\,904$/ \\ \p{24}$4\,004$\p{/}} &    $28\,228$    &          $21\,909$           & \raisebox{-.5\height}{\includegraphics[height=2em, width=\linewidth, type=pdf,ext=.pdf,read=.pdf]{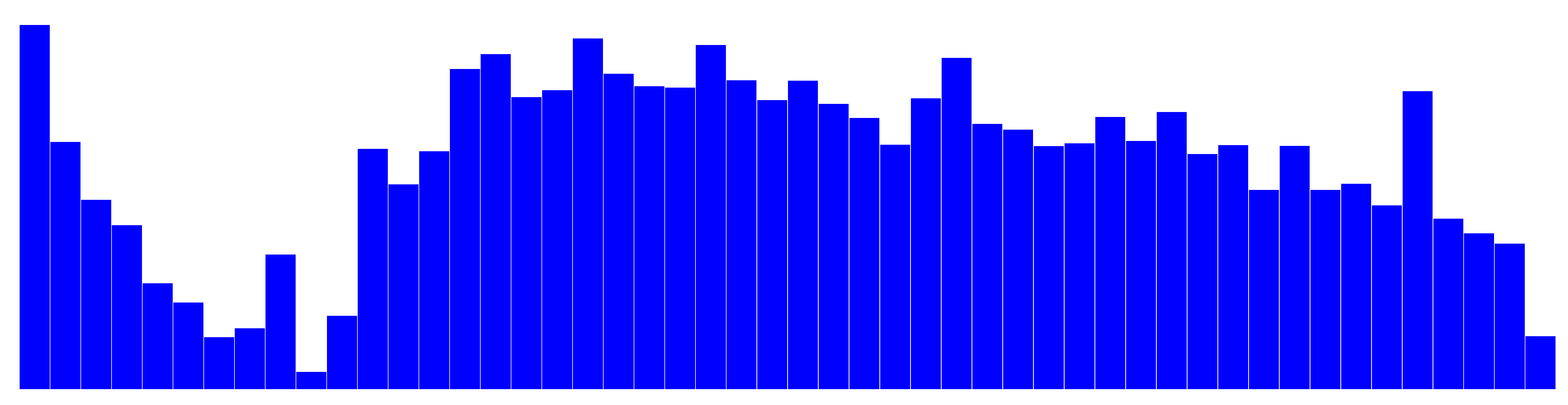}}  \\ \cline{0-3}\cline{6-6}
    2                &         $[8\,000,\infty]$          &   \makecell{\p{2}$16\,960$/ \\ $131\,736$\p{/}}   &    $24\,459$    &                              & \raisebox{-.5\height}{\includegraphics[height=2em,, width=\linewidth, type=pdf,ext=.pdf,read=.pdf]{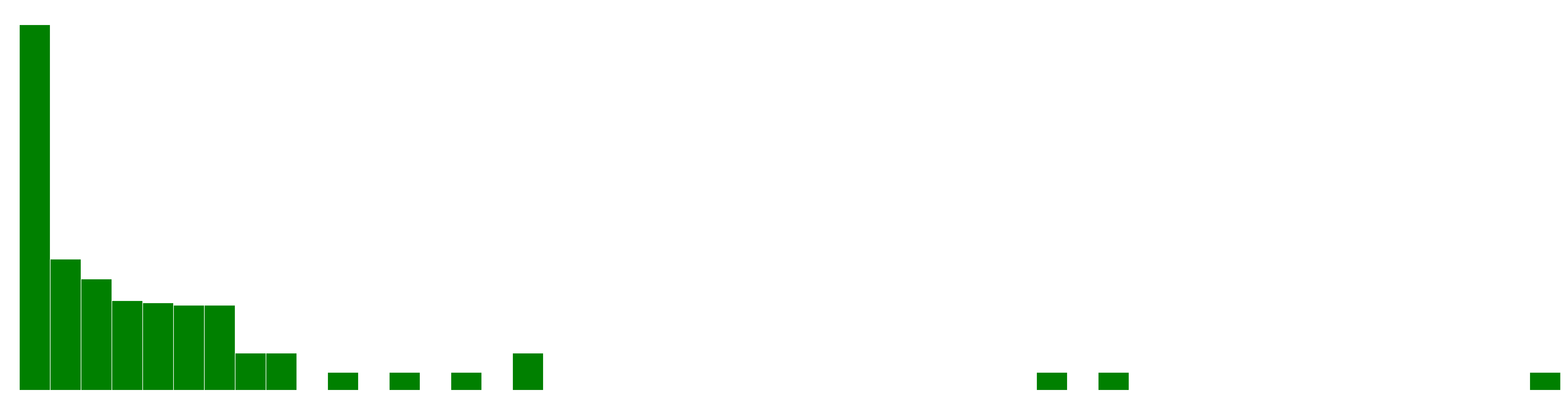}} \\ \hline
  \end{tabular}
\end{table}

\begin{figure}
  \centering
  \includegraphics[width=.9\textwidth,type=pdf, ext=.pdf, read=.pdf, trim = 0cm 0cm 0cm 0.4cm]{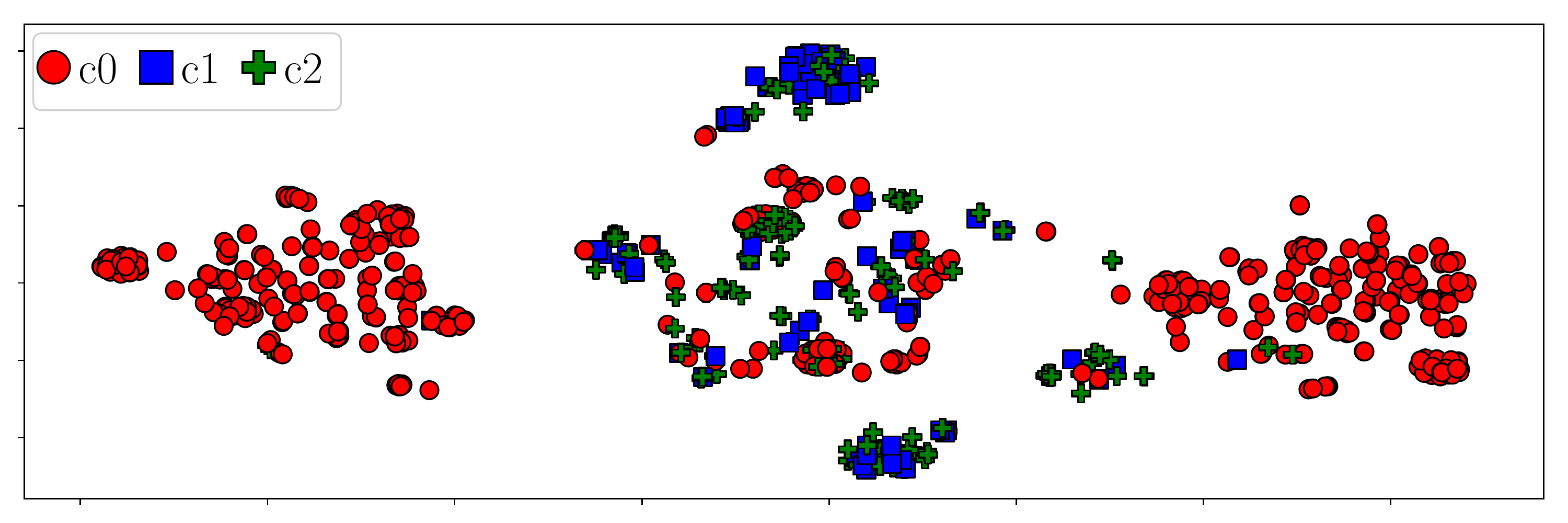}
  \caption[]{
  Visualization of the t-SNE results based on the true labels for the selected \num{1000} samples.
  Tagging is based on the three class labels ((\SI{65}[c0\approx]{\percent}, \num{652} flows, \iconl1), (\SI{12}[c1\approx]{\percent}, \num{120} flows, \iconl2), (\SI{23}[c2\approx]{\percent}, \num{228} flows, \iconl3)), whereby the exemplary boundaries are used.
  }
  \label{fig:tsne_class_labeling}
\end{figure}

\cref{fig:tsne_class_labeling} depicts the distribution of the true labels within the t-SNE output.
Whereas some spots primarily have data samples belonging to the same class (c0), other spots are a mixture of different (c1, c2) or all classes.
With regard to \cref{fig:tsne_kmeans_centers}, the results of k-means clustering cannot be used to classify the samples adequately.
Respectively, it is not sufficient to predict the bit rate of a flow.

\begin{figure}
  \centering
  \subfloat{\includegraphics[width=0.79\textwidth,type=pdf,ext=.pdf,read=.pdf, trim = .4cm .4cm .4cm .3cm, clip]{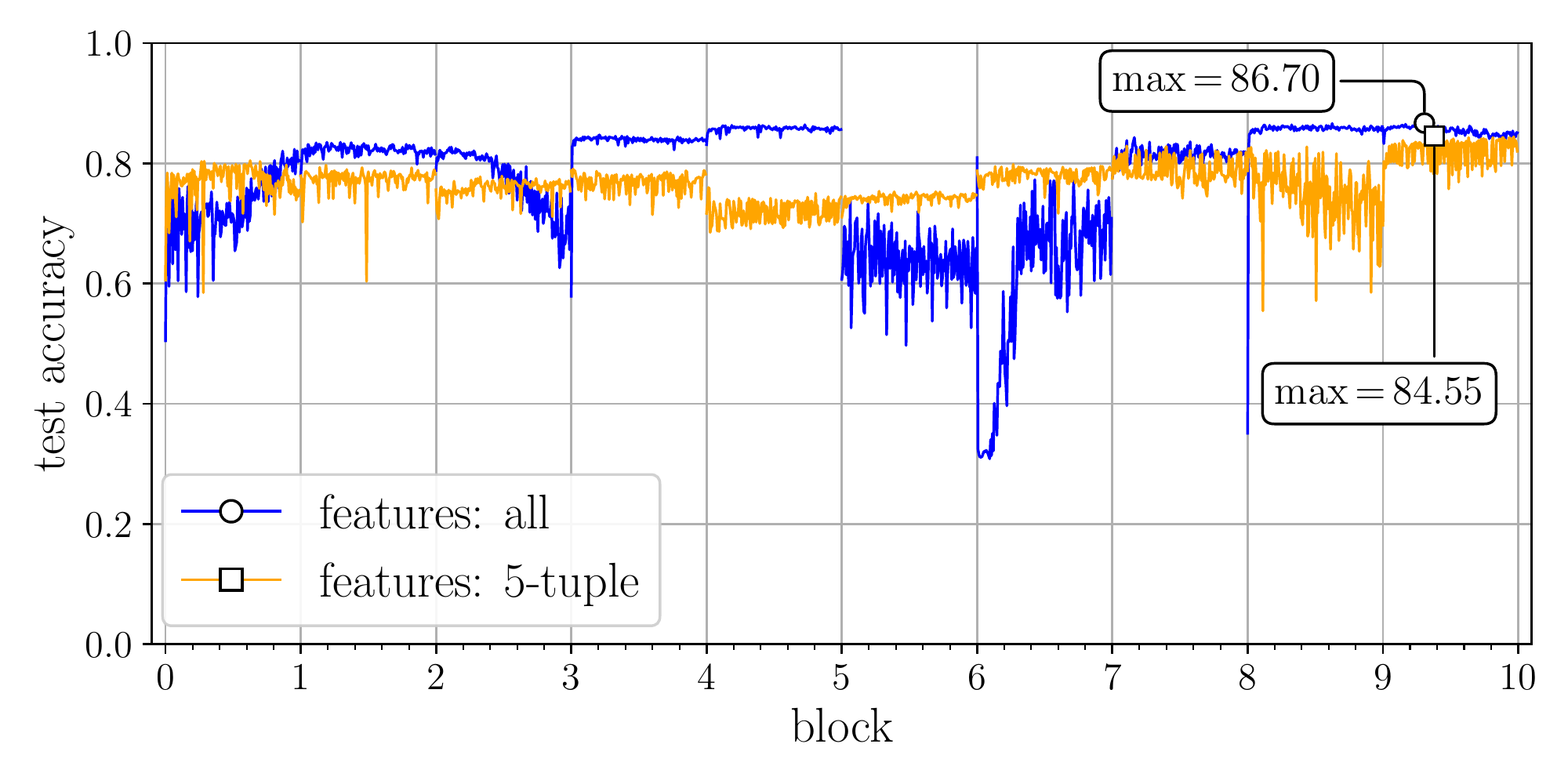}\label{fig:accuracy}} %
  \centering\Shortstack{
  \def\stacktype{L}
    \stackunder{\includegraphics[width=0.19\linewidth,type=pdf,ext=.pdf,read=.pdf, trim = 1.8cm 1.5cm 3cm .4cm, clip]{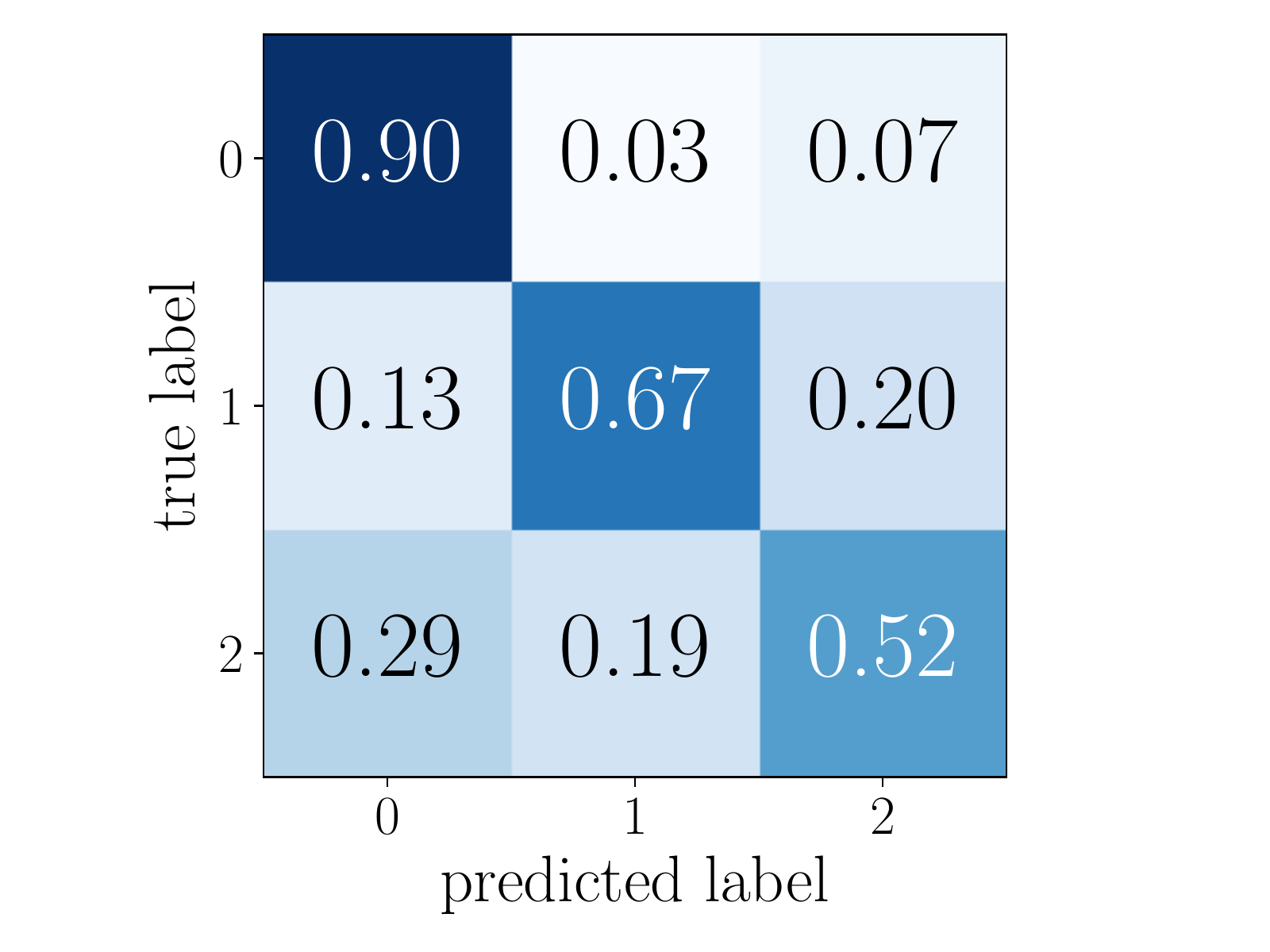}}
    \\
  \def\stacktype{L}
    \stackunder{\includegraphics[width=0.19\linewidth,type=pdf,ext=.pdf,read=.pdf, trim = 1.8cm 0cm 3cm .4cm, clip]{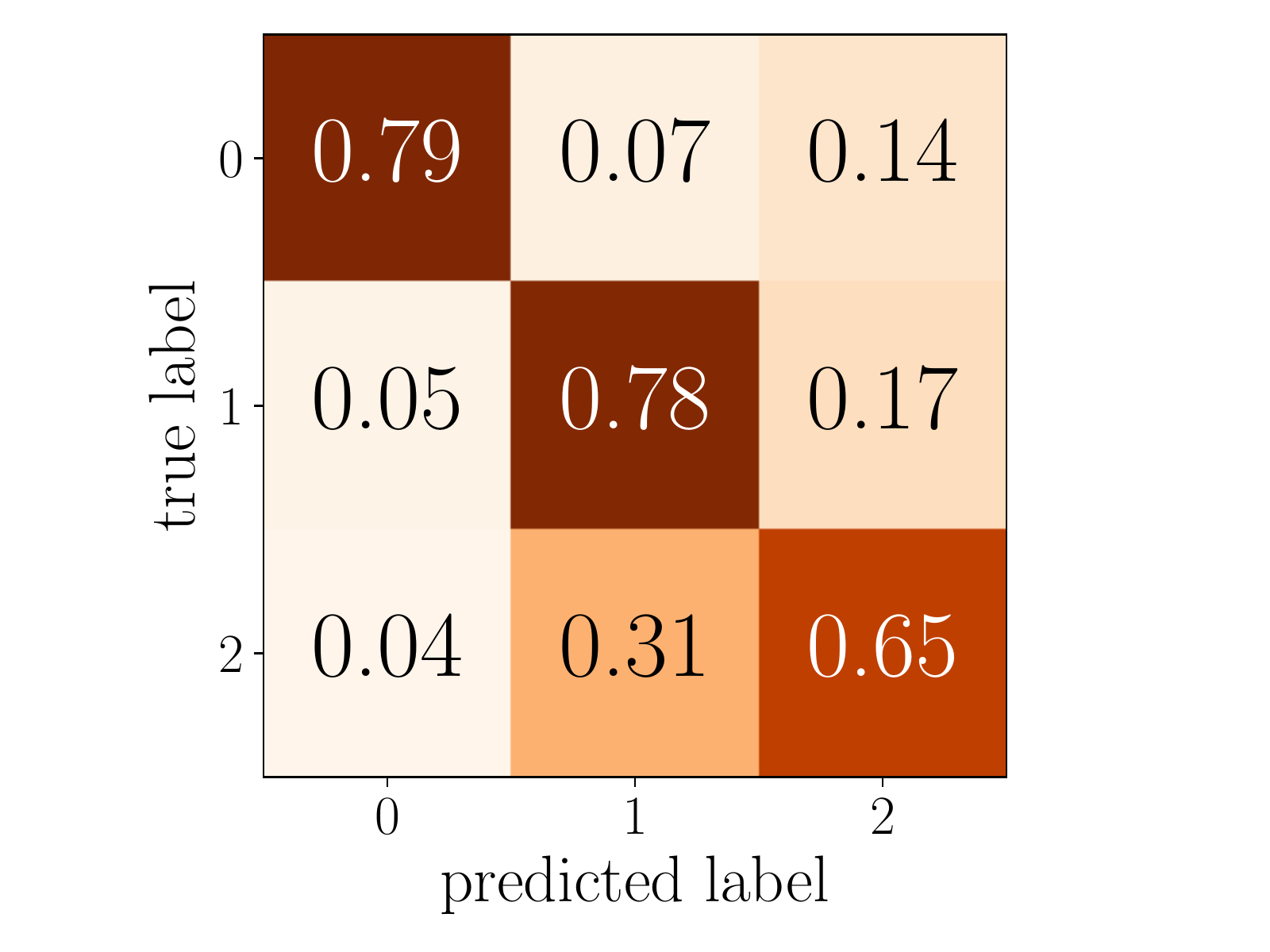}}
  } %
  \vspace{0.2cm}
  \caption[]{
    Testing results for the best experiments determined by the parameter optimization:
    first ($\mathbb{F}$\,=\,$all$, blue, \icone1) and second experiment ($\mathbb{F}$\,=\,$5\text{-}tuple$, orange, \icone2).
  The trend of the accuracy for the first \num{10} blocks is depicted.
  Training and testing is done sequentially on each independent block for \num{10} epochs.
  Besides this, the normalized confusion matrices for the last iteration of block 0 in the first (top, blue) and second experiment (bottom, orange) are shown.
  Hyper-parameters $\mathbb{F}$\,=\,$all$: $\mathcal{C}$\,=\,$[0, 50, 8\,000]$, $(d_i$,$d_h)$\,=\,$(1.0$,$1.0)$, $L$\,=\,$3$, $S$\,=\,$1000$, $\epsilon$\,=\,$0.0001$, $\mathcal{W}$\,=\,$1$, ${C}_k$\,=\,$0$;
  parameters $\mathbb{F}$\,=\,$5\text{-}tuple$: $\mathcal{C}$\,=\,$[0, 50, 8\,000]$, $(d_i$,$d_h)$\,=\,$(0.9$,$0.6)$, $L$\,=\,$5$, $S$\,=\,$1000$, $\epsilon$\,=\,$0.001$, $\mathcal{W}$\,=\,$0$, ${C}_k$\,=\,$0$.
  }
  \label{fig:accuracy_confusion_matrix}
\end{figure}

\begin{figure}[ht!]
  \centering
  \includegraphics[width=.9\textwidth,type=pdf,ext=.pdf,read=.pdf,trim = .0cm 0cm .0cm .0cm, clip]{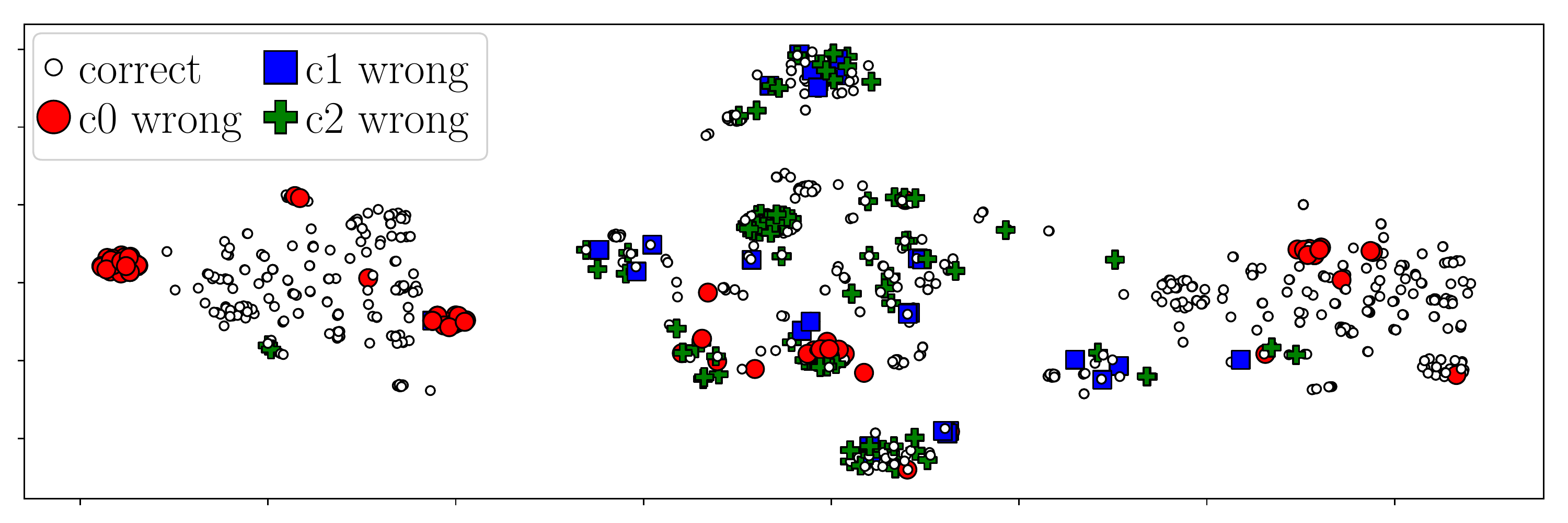}
  \caption[]{
    Visualization of the t-SNE results based on the predicted labels for the selected \num{1000} flow samples.
    Correct classified samples are marked with \iconl4 (correct\SI{77}[\approx]{\percent}, \num{775}\,flows).
    For all false classified samples ((c0 wrong\SI{9}[\approx]{\percent}, \num{86} flows, \iconl1), (c1 wrong\SI{3}[\approx]{\percent}, \num{32} flows, \iconl2) and (c2 wrong\SI{11}[\approx]{\percent}, \num{107} flows, \iconl3)) the true label is shown.
  }
  \label{fig:wrong_labels}
\end{figure}

\begin{figure}
  \centering
  \includegraphics[width=\textwidth,type=pdf,ext=.pdf,read=.pdf, trim = 0cm -.5cm 4.4cm 1.cm, clip]{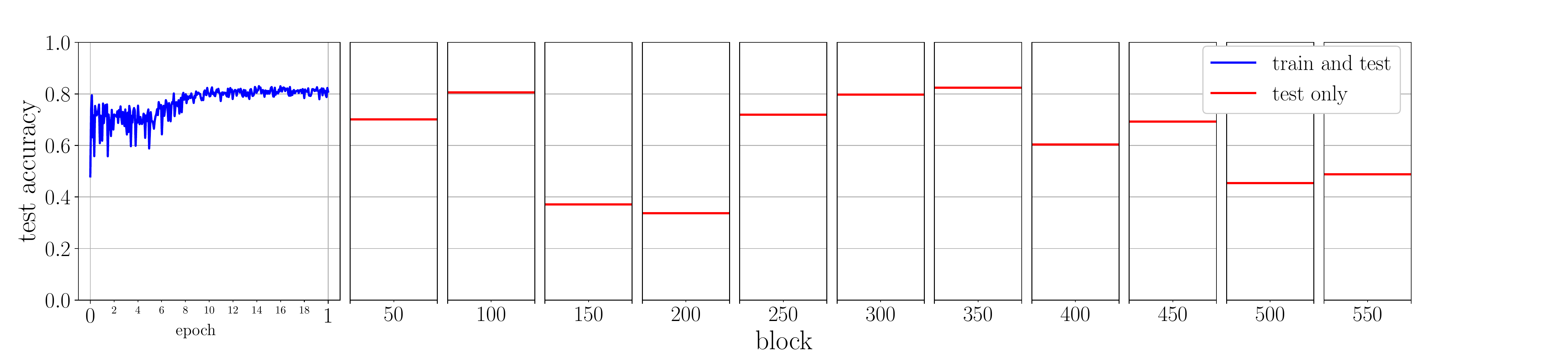}
  \caption{
    Overview of the accuracy for different blocks of the dataset.
    A DNN is trained and tested for \num{20} epochs on the first block using all available flow features (blue line).
    Subsequently, only the accuracy is determined for each fiftieth block while measuring on test data for one epoch (red lines).
    Parameter setup: see first experiment in \cref{fig:accuracy_confusion_matrix}.
  }
  \label{fig:concept_drift}
\end{figure}

The two experiments with the highest accuracy, determined by the parameter optimization, are shown in \cref{fig:accuracy_confusion_matrix}.
In the first experiment, training is done on all available flow features (247 inputs), whereas in the second one only the 5-tuple (104 inputs) is used.
\cref{fig:accuracy_confusion_matrix} depicts the trend of the prediction accuracy.
At the beginning of a directly following block, the accuracy value can considerably vary compared to the rate for the previous block but generally stabilizes for each block after a few training iterations.
This indicates a slight change in statistics (concept drift) between the individual blocks, which becomes clearer in \cref{fig:concept_drift}.
We achieve a maximum accuracy of \SI{\approx 87}{\percent} for the first respectively \SI{\approx 85}{\percent} for the second experiment.
Regarding these maxima, the data enrichment leads to an accuracy increase of about \SI{2}{\percent}.
Normalized confusion matrices for both experiments are also outlined in \cref{fig:accuracy_confusion_matrix}.
Further evaluation metrics are outlined in \cref{tab:stupidresults}.
\cref{fig:wrong_labels} gives an overview of the false classified data samples.
With regard to the false labels, prediction errors for coherent spots mainly belong to the same class.

\begin{table}
\caption{\label{tab:stupidresults}
Common binary classification measures, given separately for each of the three classes in a one-against-all setting. 
These measures are instructive, particularly when comparing performance between the two experiments (5-tuple against all features). 
The values can be computed from the unnormalized confusion matrices.
}
\centering
\begin{tabular}{|cc|c|c|c|c|}
	\hline
	\multicolumn{2}{|c|}{\textbf{Experiment}}                    & \textbf{Precision} & \textbf{Recall/Sensitivity} & \textbf{Specificity} & \textbf{Accuracy} \\ \hline
	     \multirow{3}{*}{$\mathbb{F}$\,=\,$all$}       & class 0 &        85.1        &            89.8             &         75.6         &       84.2        \\
	                                                   & class 1 &        50.0        &            66.7             &         92.0         &       89.3        \\
	                                                   & class 2 &        69.2        &            52.4             &         90.7         &       79.8        \\ \hline
	\multirow{3}{*}{$\mathbb{F}$\,=\,$5\text{-}tuple$} & class 0 &        96.6        &            79.2             &         95.7         &       85.7        \\
	                                                   & class 1 &        38.8        &            78.0             &         85.3         &       84.5        \\
	                                                   & class 2 &        64.5        &            64.8             &         85.8         &       79.8        \\ \hline
\end{tabular}
\end{table}

%
%
\section{Discussion and Principal Conclusions}\label{sec:conclusion}
The principal conclusions we can draw from the presented experiments are:
First of all, DNNs are a feasible tool for performing fine-grained network traffic flow prediction in a \enquote{big data} setting, achieving an accuracy of roughly \SI{87}{\percent} even though performed in a streaming fashion on successive and independent blocks of flow data.
Previous studies reached accuracies over \SI{90}{\percent} but grouped network flows in only two classes (\enquote{mice} and \enquote{elephant} flows), which is considerably less useful for fine-grained network traffic engineering, and, above all, processed all training data in a single block.
Secondly, we find that data enrichment can be useful, as it improves classification accuracy by roughly \SI{2}{\percent} at manageable computational cost.
Thirdly, our visualization and clustering studies show that there is no simple way to improve results by outlier detection, presumably because the data samples do not lend themselves to clustering using Euclidean distance, and a custom distance metric would have to be used here. We establish nevertheless that \mbox{t-SNE} is a useful tool to visualize structures and relations in network flow data.
Lastly, we confirm by experiments that there is moderate to strong concept drift in flow data, and that appropriate measures will have to be taken in future works to address this issue.
\par
\noindent \textbf{Comparability and validity of results}
We may ask how generalizable our results are, and the answer is of course complex.
In a university campus scenario such as ours, there are numerous factors that may affect the results, like the day of week, the season, the proximity of tests, etc.
For example, the WiFi network -- including thousands of connected students -- represents a dynamic setup that probably cannot be solved easily for a DNN because connections are unique and non-recurring (in contrast to, e.g., communications between servers).
Identifying and excluding such \enquote{difficult} flows could conceivably improve prediction accuracy and generalizability of our results.
As stated in \cref{sec:related_work}, publicly available datasets are relatively small.
Larger datasets are not accessible, probably due to privacy issues.
Even though our campus network is unique in its structure and thus results on our data do not guarantee in any way that the approach will work in other networks, the same can be said for any of the previous studies on the subject.
The only way to show generality would be to have access to several datasets of network flows of comparable size, and to perform the same experiments on all of them.
Comparing our results to other studies on the subject is further complicated by the fact that we perform three-class classification whereas previous studies were concerned with two-class scenarios only.
\par
\noindent \textbf{Discussion of the three-class scenario}
To show that our architecture can replicate previous results, we trained our DNNs on a two-class task with a threshold value of \num{500} bits per second between \enquote{mice} and \enquote{elephant} flows and obtain a test accuracy of over \SI{90}{\percent}, which is comparable to the results of other studies while taking the abovementioned caveats into consideration.
Obviously, introducing an additional class degrades the classification accuracy, simply because guessing has a lower chance of success with one more class to choose from.
Whether this lower prediction accuracy is compensated by the benefit of a more fine-grained prediction would have to be tested in simulation, which is what we are currently working on.
For this study, we wished to establish that more than two classes can be successfully integrated into a prediction scheme, all the more since the computational cost of predicting more classes is negligible at inference time.
When also considering that we perform learning in a streaming setting, which in general degrades performance w.r.t.\ settings where all data are simultaneously available for training, our results must be considered very competitive.
\par
\noindent \textbf{Justification of using DNNs}
The principal reason for using DNNs as opposed to other methods proposed in the literature, e.g., Gaussian Process Regression (GPR) \cite{poupart2016online}, is the fact that in future we want to train our classifiers in a streaming fashion: As soon as a new data block has been collected, model re-training is conducted automatically, and the trained model is immediately deployed and used for flow classification.
This puts a strong focus on the scalability of the training process w.r.t.\ the number of data samples.
In \cite{poupart2016online}, a training complexity of $\mathcal{O}(n\cdot m^2)$ is reported for GPR, where concrete values for $m$, or how they are chosen, are unclear.
Naively, GPR has a training complexity of $\mathcal{O}(n^3)$, and it is unclear whether the optimizations discussed in \cite{poupart2016online} can be tuned without human intervention (no code is provided).
In contrast, DNNs have a natural training complexity of $\mathcal{O}(n)$ without any optimizations, so they do seem a more natural choice in the \enquote{big data} context.
We will investigate the performance of other learning algorithms in future work, and compare them to our approach.
\smallskip
\par
\bigskip
\noindent \textbf{Acknowledgements}
We thank Sven Reißmann from the university data center for assistance with data collection and preparation.
We gratefully acknowledge the support of NVIDIA Corporation with the donation of the Titan Xp GPU.
\newpage
\bibliographystyle{splncs04}
\bibliography{refs}

\begin{thebibliography}{1}
\providecommand{\url}[1]{\texttt{#1}}
\providecommand{\urlprefix}{URL }
\providecommand{\doi}[1]{https://doi.org/#1}

\bibitem{benson2010network}
Benson, T., Akella, A., Maltz, D.A.: {Network Traffic Characteristics of Data
  Centers in the Wild}. In: 10th ACM SIGCOMM conference on Internet measurement
  (2010)

\bibitem{Maaten2008}
van~der Maaten, L., Hinton, G.: {Visualizing Data using t-SNE}. Journal of
  Machine Learning Research 9(Nov)  (2008)

\bibitem{nguyen2008survey}
Nguyen, T.T., Armitage, G.J.: {A Survey of Techniques for Internet Traffic
  Classification using Machine Learning}. IEEE Communications Surveys  (2008)

\bibitem{pfuelb2019a}
Pf\"ulb, B., Gepperth, A.: {A Comprehensive, Application-oriented Study of
  Catastrophic Forgetting in DNNs}. In: International Conference on Learning
  Representations ({ICLR}) (2019)

\bibitem{poupart2016online}
Poupart, P., Chen, Z., Jaini, P., Fung, F., Susanto, H., Geng, Y., Chen, L.,
  Chen, K., Jin, H.: {Online Flow Size Prediction for Improved Network
  Routing}. In: IEEE 24th International Conference on Network Protocols (ICNP)
  (2016)

\bibitem{SHI20171}
Shi, H., Li, H., Zhang, D., Cheng, C., Wu, W.: {Efficient and Robust Feature
  Extraction and Selection for Traffic Classification}. Computer Networks 119
  (2017)

\bibitem{Valadarsky:2017:LR:3152434.3152441}
Valadarsky, A., Schapira, M., Shahaf, D., Tamar, A.: {Learning to Route}. In:
  16th ACM Workshop on Hot Topics in Networks (2017)

\bibitem{wang2016framework}
Wang, P., Lin, S.C., Luo, M.: {A Framework for {QoS}-aware Traffic
  Classification using Semi-supervised Machine Learning in {SDN}s}. In: IEEE
  International Conference on Services Computing (SCC) (2016)

\bibitem{xiao2015efficient}
Xiao, P., Qu, W., Qi, H., Xu, Y., Li, Z.: {An Efficient Elephant Flow Detection
  with Cost-sensitive in {SDN}}. In: 1st International Conference on Industrial
  Networks and Intelligent Systems (INISCom) (2015)

\end{thebibliography}
The final authenticated version is available online at \url{https://doi.org/10.1007/978-3-030-30490-4_40}.
\end{document}